\documentclass[review]{elsarticle}

\usepackage{lineno,hyperref}
\modulolinenumbers[5]

\journal{Digital Signal Processing}

\usepackage{diagbox}
\usepackage{algorithm}
\usepackage{algpseudocode}
\usepackage{multirow}
\usepackage[table,xcdraw]{xcolor}
\usepackage{svg}
\usepackage{amsmath}
\usepackage{amssymb}
\usepackage{setspace}
\usepackage{amsfonts}
\usepackage{graphics}
\usepackage{url}
\usepackage[margin=30mm]{geometry}
\makeatletter
\newcommand{\vg}{\vec{g}\@ifnextchar{^}{\,}{}}
\makeatother

\begin{document}

\begin{frontmatter}

\title{Towards Real-World BCI: CCSPNet, A Compact Subject-Independent Motor Imagery Framework}

\author[First]{Mahbod Nouri\corref{contrib}}
\ead{m.nouri@ed.ac.uk}

\author[Second]{Faraz Moradi\corref{contrib}}
\ead{fmora103@uottawa.ca}

\author[Third]{Hafez Ghaemi\corref{contrib}}
\ead{hafez.ghaemi@studenti.polito.it}
 
\author[Fourth]{Ali Motie Nasrabadi\corref{cor1}}
\ead{nasrabadi@shahed.ac.ir}

\cortext[contrib]{Equal contribution}
\cortext[cor1]{Corresponding author}

\address[First]{School of Informatics, The University of Edinburgh, United Kingdom}
\address[Second]{Faculty of Engineering, University of Ottawa, Canada}
\address[Third]{Department of Control and Computer Engineering, Politecnico di Torino, Italy}
\address[Fourth]{Department of Biomedical Engineering, Shahed University, Tehran, Iran}

\begin{abstract}
A conventional brain-computer interface (BCI) requires a complete data gathering, training, and calibration phase for each user before it can be used. In recent years, a number of subject-independent (SI) BCIs have been developed. Many of these methods yield a weaker performance compared to the subject-dependent (SD) approach, and some are computationally expensive. A potential real-world application would greatly benefit from a more accurate, compact, and computationally efficient subject-independent BCI. In this work, we propose a novel subject-independent BCI framework, named CCSPNet (Convolutional Common Spatial Pattern Network) that is trained on the motor imagery (MI) paradigm of a large-scale electroencephalography (EEG) signals database consisting of 400 trials for every 54 subjects who perform two-class hand-movement MI tasks. The proposed framework applies a wavelet kernel convolutional neural network (WKCNN) and a temporal convolutional neural network (TCNN) in order to represent and extract the spectral features of EEG signals. A common spatial pattern (CSP) algorithm is implemented for spatial feature extraction, and the number of CSP features is reduced by a dense neural network. Finally, the class label is determined by a linear discriminant analysis (LDA) classifier. The CCSPNet evaluation results show that it is possible to have a compact BCI that achieves both SD and SI state-of-the-art performance comparable to complex and computationally expensive models.
\end{abstract}


\end{frontmatter}


\section{Introduction} \label{intro}
Brain-computer interface (BCI) is a communication system that exploits several methods to decode, interpret, and classify intentions and decisions using neural activities \cite{leuthardt2004brain,schalk2004bci2000}. One of the most prevalent types of BCI uses recorded brain signals, known as electroencephalography (EEG). There are generally three different paradigms in EEG-based BCI \cite{lee2019eeg}. These paradigms include motor imagery (MI) \cite{pfurtscheller2001motor}, event-related potential (ERP) \cite{wu2016novel}, and steady-state visually evoked potential (SSVEP) \cite{friman2007multiple}.

The BCI paradigm with the most potential in real-world applications, especially for rehabilitation engineering, is motor imagery. Motor imagery can be defined as a mental process in which a person imagines performing a particular action without any physical movement \cite{decety1990brain,neuper2005imagery}. The applications imaginable for MI-BCI are unlimited. MI is useful for powering equipment designed to help disabled people execute specific actions they could not have performed otherwise due to their disability \cite{bright2016eeg,jacob2019artificial}. Furthermore, motor imagery is even set to transform the world of entertainment in the future \cite{li2017development,bordoloi2012motor}. In this paper, we will focus our attention on this paradigm.

Most of the works done in the MI domain are subject-dependent; i.e., the proposed methods require data recording, training, and calibration for individual subjects. This calibration phase is necessary for a reliable decoder specific to each subject \cite{lotte2007review}. This subject specificity may hinder the use of BCI in real-world applications. In a real-world BCI application, e.g., portable wearable devices or smart wheelchairs \cite{wang2018wearable,li2013design}, the EEG signals may be recorded with mobile headwear and EEG caps that are now becoming widely available, and the device may utilize low-power edge-computing to predict the subject’s decision in real-time \cite{rajesh2019secure}. This type of application not only demands a subject-independent framework, it also would benefit from compactness and memory efficiency. Having in mind both the subject-dependency problem and the need for compactness and efficiency, we are proposing a hybrid deep learning framework, named CCSPNet, for subject-independent MI decoding.

The main contributions of this framework are given below:

\begin{enumerate}
\item We employed a dynamic digital signal filtering technique for spectral feature extraction based on convolutional neural networks, utilizing a combination of smooth and non-smooth filter kernels. We hypothesize that this dynamic combination may be able to extract both subject-specific features and generalizable subject-independent ones that could improve the overall SI performance of the framework. These filters are optimized to produce better features for the common spatial pattern algorithm.
\item We integrated common spatial pattern (CSP) for spatial feature extraction into a deep learning framework, and used a novel loss function based on the quality of the CSP features to update the weights of the digital filters that generate CSP inputs. This feedback mechanism for optimizing CSP inputs has enabled our framework to outperform traditional CSP methods.
\item The framework without the classification stage may be used as a compact, yet powerful spatio-temporal feature extractor for a more complex decoder in other BCI models.
\end{enumerate}

The subsequent sections of this article are organized as follows:

In section~\ref{rel}, we will discuss some of the earlier efforts and related works in motor-imagery decoding. Section~\ref{data} will be dedicated to describing the dataset used in our work. In section~\ref{method}, CCSPNet and its architecture will be described. The evaluation results and a discussion around the characteristics of the proposed framework will be given in section~\ref{res}. A conclusion is offered in section~\ref{con}.

\section{Related Works} \label{rel}
The majority of research works in the BCI literature are concerned with the subject-dependent (SD) approach. As mentioned before, SD models require a training and calibration phase to adapt to each subject’s brain patterns. Some of these models have used conventional source analysis \cite{qin2004motor}, or machine learning (ML) algorithms for classification. Wang et al. \cite{wang2006common} used common spatial patterns (CSP) for channel selection and linear discriminant analysis (LDA) for classification. Dong et al. \cite{dong2017classification} proposed a hierarchical method for extending Support Vector Machines (SVM) to be used on a 4-class BCI dataset. They used one-versus-rest CSP and one-versus-one CSP in each of their SVM layers. 

Recently, the utilization of deep learning (DL) algorithms for decoding EEG signals and task classification has become a trend in the BCI literature\cite{sors2018convolutional,al2021deep,sarkar2022deep,llorella2021classification}. Lawhern et al. have proposed a well-known compact convolutional neural network (CNN), known as EEGNet\cite{lawhern2018eegnet}, that has been widely applied to various datasets and BCI paradigms for task classification, and, in many cases, yields an acceptable subject-dependent accuracy. There are models inspired by EEGNet, which are also compact and suitable for low-power computing \cite{huang2020s,schneider2020q}. 

Time-frequency analysis techniques on EEG signals have been employed to pre-process neural network inputs. For example, Zhang et al. \cite{zhang2019novel} applied complex Morlet wavelet transforms on EEG data to feed their convolutional neural network (CNN) with a time-frequency tensor. Molla et al. \cite{molla2021trial} used multivariate discrete wavelet transforms to decompose EEG channels data into sub-bands, which were then processed using a CSP filter to extract features. Other time-frequency analysis methods, such as short-time Fourier transform (STFT), have also been adopted to create a more suitable input for CNNs \cite{shovon2019classification,tabar2016novel}. 

One of the main challenges facing DL-BCI researchers has been the scarcity of large-scale datasets (regarding both the number of subjects and the number of trials per subject) to be utilized in deep learning algorithms. A possible solution to this problem is data augmentation. Traditional data augmentation approaches such as empirical mode decomposition \cite{zhang2019novel,dinares2018new} and geometric transforms such as rotation, flipping, and zooming on STFT images \cite{shovon2019classification} have been used in the BCI literature. Novel approaches such as data augmentation using generative adversarial networks (GANs \cite{goodfellow2014generative}) have also been implemented to achieve higher accuracies with neural networks in subject-dependent MI \cite{fahimi2020generative,zhang2018improving}. 

The works discussed up to now have all been subject-dependent and require separate training and tuning for each subject. In order to reduce the training time for a new subject, it is possible to use transfer learning from models trained on other subjects \cite{fahimi2019inter,sakhavi2017convolutional,jayaram2016transfer,azab2019weighted}. Still, transfer learning is not the final solution to subject dependence.

A lower SI performance compared to SD in some of the other works \cite{lotte2009comparison,reuderink2011subject} may be attributed to the fact that spectral-spatial patterns of different subjects' brain signals vary significantly, and traditional machine learning algorithms have a limited capacity for complex pattern recognition. Deep learning and neural networks have the unique ability to adapt to highly variable data and detect complex patterns, and therefore, can be effectively used in the SI BCI approach. However, deep neural networks require a large number of trials (training samples) for different subjects to achieve a good subject-independent performance. One of the first frameworks that has taken advantage of a large-scale MI database to implement a deep learning framework for motor imagery paradigm, is proposed by Kwon et al \cite{kwon2019subject}. They used a CNN-based framework with around 72 million free parameters and have been able to achieve a SI performance that is comparable to their SD performance. In this framework, the parallel CNNs process the CSP features extracted from different frequency bins, selected based on an information-theoretic analysis. A concatenation fusion layer and a fully-connected layer are then used to output the predicted class labels. In another work, Autthasan et al. \cite{autthasan2021min2net} used a multi-task autoencoder architecture to achieve a latent representation of the EEG data with lower dimensionality. As part of the training process, a supervised classifier and deep metric learning were applied to the latent layer to enhance class discrimination.

The CCSPNet framework, proposed in this work, achieves state-of-the-art subject-independent performance while maintaining a low number of model parameters and remaining compact. 

\section{Dataset} \label{data}
The dataset used to evaluate the performance of our framework is an open-source BCI database recorded by Lee et al. \cite{lee2019eeg}. This database is one of the largest BCI datasets recorded to this date. 54 healthy subjects, including 25 females and 29 males, aged 24 to 35, went through the same experiment in two sessions, among whom were 16 experienced BCI users and 34 inexperienced ones. 62 Ag/AgCl electrodes were used to capture EEG signals with a sampling rate of 1000 Hz. Furthermore, four electrodes recorded EMG signals for each individual. The experiment, including all three BCI paradigms (SSVEP, ERP, and MI), was conducted in two similar sessions.

\begin{figure}[h]
\begin{center}
    \centering
	\includegraphics[scale=0.2]{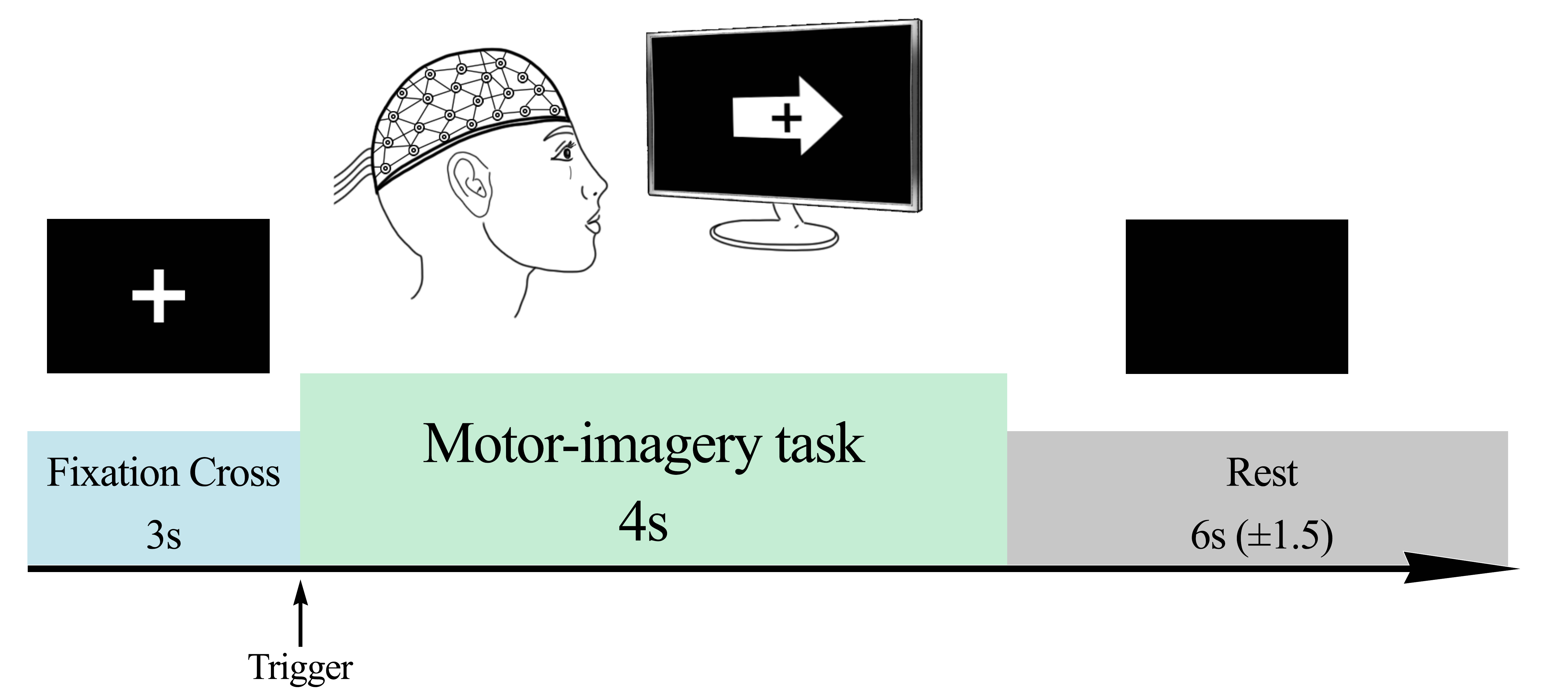}
	\caption{The binary MI paradigm EEG signal recording}
	\label{fig1}
\end{center}
\end{figure}

Our focus is on the motor imagery paradigm task classification, and we will further discuss the MI experiment. Each MI session is composed of an offline and an online phase, each including 100 trials. During each offline trial, subjects were prompted to execute a binary motor imagery task. The timing of a single MI trial in this phase is illustrated in Figure~\ref{fig1}. At the beginning of the trial, as preparation, a black fixation cross appeared on the screen center for three seconds. Afterward, a prompt instructing the subjects to perform the MI task was shown. The prompt was a left or right arrow standing for the movement imagination of the left or the right hand. After each trial, a period of $6\pm1.5$ seconds blank screen was shown, indicating the rest period.

When the offline phase ended, a set of classifiers based on common spatial pattern (CSP) and linear discriminant analysis (LDA) were trained using the recorded EEG signals. In the online phase, subjects perform the MI tasks, and their performance is evaluated using the trained online classifiers. Real-time neurofeedback in the form of a right or left arrow appeared on the screen, denoting the predicted class. This feedback is designed to improve the user performance in future trials.

\section{Methodology} \label{method}
The framework proposed in this article for EEG decoding is designed in a modular manner. We can identify three blocks in CCSPNet:

\begin{itemize}
    \item \textbf{Temporal feature extraction}: This block consists of one-dimensional CNNs. A wavelet kernel convolutional neural network (WKCNN) is used for dynamic smooth filtering, and a temporal convolutional neural network (TCNN) for dynamic non-smooth filtering.
    \item \textbf{Spatial feature extraction}: This part consists of a common spatial pattern algorithm for spatial feature extraction followed by a dense artificial neural network (ANN) for feature reduction.
    \item \textbf{Classifier}: The last stage is a classifier that determines the predicted class. We used an LDA classifier in this block.
\end{itemize}

\subsection{Pre-processing}
The raw EEG data for a single trial are denoted by a matrix of dimensions $(C, T)$ where $C$ is the number of channels (electrodes), and $T$ is the number of sample points. In the dataset used for evaluation, $C = 62$, and $T = 4000$. All training trials can be concatenated into a 3-D matrix designated as $X$, and with dimensions $(N, C, T)$. Similarly, the Label vector is constructed by concatenating all trial labels.

As the first pre-processing step, the EEG signals were trimmed, and the time points between 1000 to 3500 milliseconds were extracted. Motor imagery tasks are generally associated with the $\mu$ (7.5-12.5 Hz) and $\beta$ (12.5-30 Hz) frequency bands. Specifically, it is shown that performing MI tasks causes an energy decrease in the $\mu$ band (event-related desynchronization or ERD) and an increase in the beta band (event related synchronization or ERS). MI tasks related to the left and right hand movements cause ERD and ERS in the right and left sides of the motor cortex, respectively \cite{pfurtscheller1999event}. We down-sampled the signals to 100 Hz to ensure that the information from the related frequency bands is preserved.

EEG signals related to motor imagery tasks are mostly generated in the motor cortex. In the pre-processing stage, data recorded by channels other than those traditionally associated with their relative BCI paradigm, in this case, MI may be removed \cite{zhang2019novel,kwon2019subject} for reasons such as reducing input dimensionality, model complexity, training time, and overfitting \cite{alotaiby2015review}. However, recorded EEG signals do not have a perfect spatial resolution, i.e., the channels may capture information from areas of the brain other than those in their vicinity. Furthermore, in many applications, we may not have a priori knowledge about the channels, and as argued by Lawhern et al. \cite{lawhern2018eegnet}, deep learning models that do not extract information from specific channels provide a middle ground between input dimensionality and the flexibility to extract all relevant features from the data. In our model, we follow the second approach and do not decrease the number of channels.

As the second pre-processing step, a fifth-order band-pass Butterworth filter, $f$, with the conventional frequency band, $[8,30 Hz]$, was applied to the data:

\begin{equation} \label{eq1}
X_f = f \circledast X
\end{equation}

where $\circledast$ denotes filtering operation. This frequency range is a conventional one for classifying hand-movement MI signals that encompasses the alpha and beta EEG frequency bands that have been shown to be the most important bands in such a task \cite{ramoser2000optimal,pfurtscheller1997eeg}, which also achieves the best performance in our framework. We also performed classification by including frequencies from delta, theta, and gamma bands, however the $[8,30 Hz]$ band outperformed the rest, and only the results related to this band are reported.

\subsection{Proposed Method}
\begin{figure}[h]
    \begin{center}
    \centering
	\includegraphics[scale = .17]{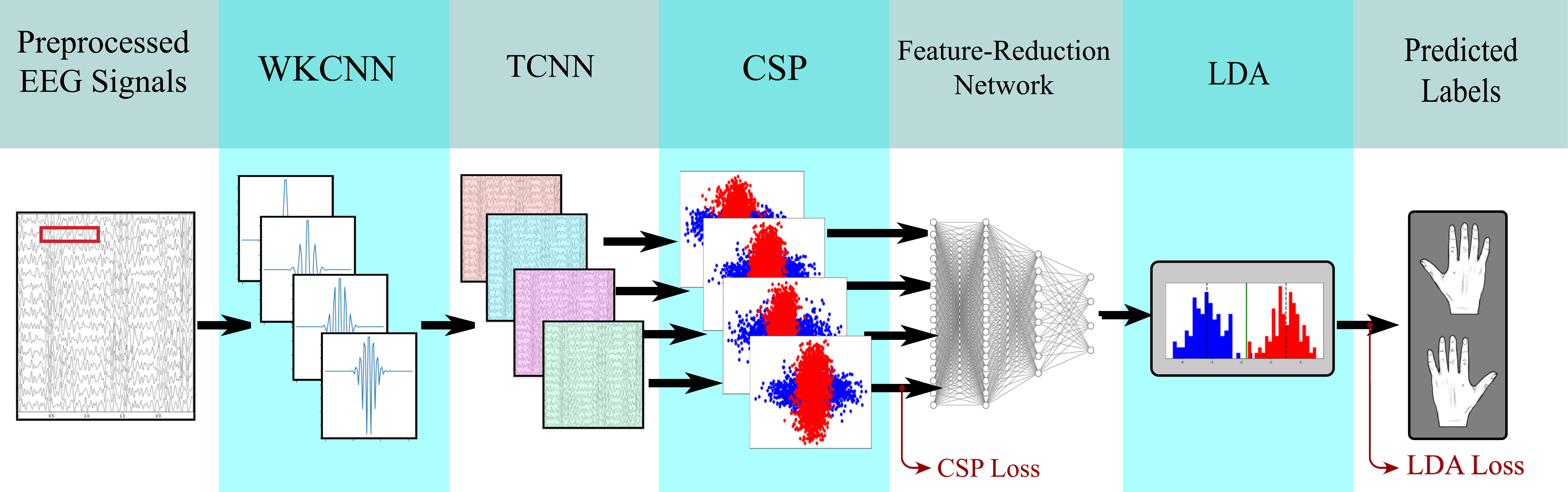}
	\caption{Graphical representation of the proposed framework (CCSPNet); The decoding algorithm starts with a wavelet kernel convolutional neural network acting upon the pre-processed EEG signals with one-dimensional wavelet kernels (first and second columns). Afterward, a temporal convolutional neural network with normal one-dimensional convolutional kernels will be applied to the data (third column). These two networks are spectral feature extractors and do not reduce the number of channels. Then, a common spatial pattern algorithm will be applied to each CNN output to filter the data spatially and extract spatial features (fourth column), and a dense artificial neural network is used for CSP feature reduction (fifth column). Finally, an LDA classifier outputs the predicted class labels (sixth and seventh columns).}
	\label{fig2}
	\end{center}
\end{figure}

A graphical representation of the proposed framework is presented in Figure~\ref{fig2}. A summary of the architecture is also given in Table~\ref{tab1}. The following paragraphs will describe the decoding algorithm in detail.

\begin{table}[h]
\begin{center}
\centering
\caption{CCSPNet Architecture Summary}
\begin{tabular}{cccc}
\textbf{Module}                                               &\textbf{Layer}    & \textbf{Output Dimension} \\ 
\hline
\rowcolor[HTML]{F2F2F2} 
-                                                              & Input            &(N,1,64,250)     \\
Wavelet-kernel CNN                                             & Convolution with (1, 64) kernels      &(N,4,64,250)      \\
\rowcolor[HTML]{F2F2F2} 
Wavelet-kernel CNN                                                   & Batch Norm 2D     &(N,4,64,250)      \\
Temporal CNN                                             & Convolution with (1, 32) kernels      &(N,4,64,250)      \\
\rowcolor[HTML]{F2F2F2} 
Temporal CNN                                                   & Batch Norm 2D     &(N,4,64,250)      \\
CSP                                                            & -                &(N,4,4)            \\
\rowcolor[HTML]{F2F2F2} 
\cellcolor[HTML]{F2F2F2}                                       & Flatten          &(N,16)             \\
\cellcolor[HTML]{F2F2F2}                                       & Dense            &(N,16)            \\
\rowcolor[HTML]{F2F2F2} 
\cellcolor[HTML]{F2F2F2}                                       & Batch Norm 1D    &(N,16)            \\
\cellcolor[HTML]{F2F2F2}                                       & Dense            &(N,8)            \\
\rowcolor[HTML]{F2F2F2} 
\cellcolor[HTML]{F2F2F2}                                       & Batch Norm 1D    &(N,8)            \\
\multirow{-6}{*}{\cellcolor[HTML]{F2F2F2}Dense Neural Network} & Dense            &(N,4)             \\
\rowcolor[HTML]{F2F2F2} 
LDA (Classifier)                                               & -                &(N,1)            
\end{tabular}
\label{tab1}
\end{center}
\end{table}

\subsubsection{Wavelet Kernel Convolutional Neural Network (WKCNN)}
As the first stage of the algorithm, we employed a convolutional layer consisting of four one-dimensional wavelet kernels with trainable parameters and non-linear activation functions \cite{li2021waveletkernelnet}. The wavelets used are real-valued Morlet wavelets, with minimum and maximum frequencies of 8 and 30 Hz which are consistent with the Butterworth filter range used during the pre-processing. There are three trainable parameters associated with this layer. The wavelet frequencies, the Gaussian signal’s full width at half maximum, and the Gaussian exponent coefficient for adjusting the wavelet magnitude are these trainable parameters. The real-valued Morlet wavelets consist of the multiplication of a cosine wave and a Gaussian wave as below:

\begin{equation} \label{eq2}
w=\cos (2 \pi f t) \times e^{-\frac{c t^{2}}{h^{2}}}
\end{equation}

\noindent Where $t$ denotes the time-vector constructed as follows:

\begin{equation} \label{eq3}
t=\frac{k_{w}}{f_{s}}
\end{equation}

\noindent where $c$ is the Gaussian exponent coefficient for adjusting the wavelet magnitude, $f$ is the wavelet frequency, $t$ is the time vector by which the wavelet is constructed, $k_w$ is the wavelet kernel length, $f_s$ is the sampling frequency after down-sampling the data (100 Hz), and $h$ is the Gaussian signal’s full width at half maximum. The trainable parameters are $f$, $h$, and $c$, and the wavelet frequencies are initialized linearly between 8 and 30 Hz.


\subsubsection{Temporal Convolutional Neural Network (TCNN)}
As illustrated in Figure~\ref{fig2}, the second step for temporal feature extraction is a CNN consisting of temporal kernels. In this part, dynamic kernels will be applied to the data. The nomenclature "temporal convolution" is used by Lawhern et al. \cite{lawhern2018eegnet} to emphasize that this type of convolution acts only upon the signal time points and does not reduce the number of channels.



\subsubsection{Common Spatial Pattern (CSP)} \label{csp}
After spectral feature extraction, we employ a common spatial pattern (CSP) algorithm \cite{guger2000real} to extract spatial features from the signals. Let us separate the output of the TCNN layer to form a set of four matrices with dimensions $(62, 250)$ for each CNN kernel output.
A CSP algorithm was then performed on each element of this set, and a set of projection matrices were found by solving a generalized eigenvalue problem for each element.
Because the first and last columns in a CSP projection matrix are spatial filters that would maximize the variance of one class and minimize the variance of the other in a binary MI classification task, we reduce the projections in $P$ by choosing the first two and the last two columns and construct transformation matrices by concatenating these columns together.
By performing a matrix multiplication between each element in this spatial transformation and its corresponding element in output matrix of the TCNN, we will obtain the following spatially filtered signals:
\begin{equation} \label{eq9}
\left\{E_{i}\right\}_{i=1}^{4}=\left\{W_{i}^{T} \mathrm{X}_{i}\right\}_{i=1}^4
\end{equation}

\noindent where $W_i$ is a CSP transformation matrix, $X_i$ denotes matrix i of TCNN outputs. Consequently, each $E_{i}$ has dimensions $(4, 250)$.

To extract features from these signals, we calculate the logarithm of their variances to form the following feature vectors:

\begin{equation} \label{eq10}
\{\vec v_{i}\}_{i=1}^{4} = \{\log(\sigma_{E_i})\}_{i=1}^{4}
\end{equation}

\noindent where each $\vec v_{i}$ is a feature vector of size four and $\sigma_{E_i}$ denotes the variance of $E_i$ .

In this step, we define our first loss function based on cross-entropy using the extracted feature vectors and the ground truth label. We define a target vector of size four where the first two indices are the ground truth label (zero or one), and the last two are the opposite class label. We denote this target vector with $\vec{y}_n$ for the $n^{th}$ trial. In order to compute the loss, the feature vectors must be normalized to the range $[0,1]$. Therefore, a softmax operator will be applied to each feature vector before calculating the loss.
The final loss for the $n^{th}$ trial is the sum over the loss of the four feature vectors and can be computed as follows:

\begin{equation} \label{eq12}
L_n = \sum_{i=1}^{4}\bigg(\vec{y}_n\cdot\log(\sigma(\vec{v}_{i_{norm}}))+(\vec{1}-\vec{y}_n)\cdot \log(\vec{1}-\sigma(\vec{v}_{i_{norm}}))\bigg)
\end{equation}

\noindent where $\sigma$ is the softmax operator and $a\cdot b$ indicates the internal product between vectors $a$ and $b$.

The purpose of CSP is to separate the two classes as much as possible, and $\vec{y}_n$ which is a vector normal to a hypothetical hyperplane representing the ground truth class is a reasonable choice for the target vector. By minimizing the loss in \eqref{eq12}, we are trying to minimize the distance between the CSP feature vectors and this optimal target vector to enhance the distinguishing ability of the CSP.

We use this loss function for optimizing the weights of the two CNNs used for spectral filtering in the first and second stages of the framework. In other words, this loss is providing feedback about the quality of the CSP inputs to the CNNs and can help them produce better inputs for CSP.

\subsubsection{Dense Neural Network}
Before the final classification, we use a dense neural network for optimizing and reducing the number of spatio-temporal features extracted. The features extracted in the CSP stage were integrated into a single feature vector of size 16:

\begin{equation} \label{eq14}
\vec{v}=\text {concat}\left\{v_{i}\right\}_{i=1}^{4}
\end{equation}

The artificial neural network (ANN) in this stage consists of three dense layers with 16,8, and 4 neurons, respectively.
\subsubsection{Linear Discriminant Analysis (LDA)}
In the last step of the algorithm, we use a linear discriminant analysis (LDA) classifier to determine the predicted class label. LDA has been traditionally successful in decoding CSP features \cite{wu2013common}. Additionally, in this particular dataset, during the online phase of the experiment, an LDA classifier was trained for each subject, and its decoding results were given to the subject as neurofeedback. This neurofeedback might have adapted the subject’s brain patterns to LDA over time which may improve the classification performance in our model. In the training phase, the LDA receives batched inputs from the neural network, and by solving a generalized eigenvalue problem finds a transformation, $W_l$, that can be applied to the input batch of size $b$ and compute a single number for each sample:

\begin{equation} \label{eq16}
\vg^{b}=W_{l}^{T} X_l 
\end{equation}

\noindent where $\vg^{b}$ is the output of the LDA for a batch size of $b$, and $X_l$ is a batch of feature vectors from the ANN.

The LDA classifier minimizes the Fisher's criterion defined below:

\begin{equation} \label{eq17}
J(\vg^{b})=\frac{\sigma^2_{\vg^{b}_0}+\sigma^2_{\vg^{b}_1}}{(\mu_{\vg^{b}_0}-\mu_{\vg^{b}_1})^2}
\end{equation}

\noindent where $\vg^{b}_0$ consists of indices in $\vg^{b}$ whose ground truth labels are zero, and $\vg^{b}_1$ consists of indices whose ground truth labels are one. Also, $\sigma^2_{\vec{x}}$ and $\mu_{\vec{x}}$ denote variance and mean of elements of vector $\vec{v}$, respectively. For classifying a new sample, LDA chooses the class with the minimum distance to the mean of its feature vectors.

To update the weights of the ANN used for feature reduction, we use the weighted average of the loss function in \eqref{eq12} and the Fisher's criterion-related to LDA:

\begin{equation} \label{eq18}
Loss = r \times L + (1-r) \times J
\end{equation}

\noindent where L is the cumulative loss in \eqref{eq12} for a batch of samples, J is the Fisher's criterion for that batch, and r is the loss ratio. After performing a grid search, we set $r=0.3$.

The total number of free parameters in the framework (including CSP and LDA parameters) is 5036. Algorithm \ref{alg1} outlines the complete classification procedure. 

\begin{algorithm}
\caption{CCSPNet Classification Algorithm}
\begin{algorithmic}
\item \textbf{Input: }a batch of training samples
\item \begin{itemize}
    \item { $\boldsymbol{b}$: Batch size} 
    \item { $\boldsymbol{x}\boldsymbol{\mathrm{\in }}{\mathbb{R}}^{\boldsymbol{C}\boldsymbol{\mathrm{\times }}\boldsymbol{T}}$: EEG data from a single trial with $C$ channels, and $T$ time points }
    \item { $\boldsymbol{X}\boldsymbol{\mathrm{=}}{\left\{{\boldsymbol{x}}_{\boldsymbol{i}}\right\}}^{\boldsymbol{\mathrm{b}}}_{\boldsymbol{i}\boldsymbol{\mathrm{=}}\boldsymbol{\mathrm{1}}}$: A batch of EEG training samples}
    \item{ ${\boldsymbol{Label}}\boldsymbol{\mathrm{=}}{\left\{{\boldsymbol{label}}_{\boldsymbol{i}}\right\}}^{\boldsymbol{\mathrm{b}}}_{\boldsymbol{i}\boldsymbol{\mathrm{=}}\boldsymbol{\mathrm{1}}}$: A batch of ground truth labels corresponding to $X$}
    \item{ $\boldsymbol{f}$: Fifth-order $\boldsymbol{\mathrm{[}}\boldsymbol{\mathrm{8}}\boldsymbol{\mathrm{,\ }}\boldsymbol{\mathrm{30}}\boldsymbol{\mathrm{\ }}\boldsymbol{Hz}\boldsymbol{\mathrm{]}}$ Butterworth filter}
\end{itemize}
\\
\item \textbf{Output: }Predicted labels \\

\item \textbf{Operators: }
\item \begin{itemize}
    \item{$\boldsymbol{\mathrm{\circledast }}$: The fifth-order Butterworth band-pass filter operator}
    \item{ $\boldsymbol{WK}$: The wavelet kernel convolutional neural network operator.} 
    \item{ $\boldsymbol{TC}$: The temporal convolutional neural network operator.} 
    \item{ $\boldsymbol{gep}$: Operator for solving a generalized eigenvalue problem}
    \item{ $\boldsymbol{\sigma}$: Softmax operator}
    \item{ $\boldsymbol{FR}$: The feature reduction dense neural network operator} 
    \item{ $\boldsymbol{LDA}$: Linear discriminant analysis classifier}
\end{itemize} \\

\item \textbf{Procedure: }

\State $X_f = f \circledast X$ \algorithmiccomment{pre-processing}

\State $X_{wk}=WK(X_f)$ \algorithmiccomment{WKNN as the first stage of spectral filtering}

\State $X_{tc}=TC(X_{wk})$ \algorithmiccomment{TCNN as the second stage of spectral filtering}

\State $X_{tc} \rightarrow \{X_{{tc}_i}^{62 \times 250}\}_{i=1}^4$ \algorithmiccomment{Separating TCNN outputs to form CSP inputs}

\For{$i=1$ to $4$}

\State$W_{i}^{62 \times 62}=gep(X_{{tc}_i}, Label)$ \algorithmiccomment{Computing the projection matrix}

\State$W_{r_i}^{62 \times 4} = concat\{W_{i}[:,1:2], W_{i}[:,63:64] \}$ \algorithmiccomment{Constructing the transformation matrix}

\State$E_{i}=W_{r_i}^{T} X_{{tc}_i}$ \algorithmiccomment{Spatially filtered signals}

\State$\vec{v}_{i}=\log(var(E_i))$ \algorithmiccomment{Extracting spatio-temporal features}

\State$\vec{v}_{i_{norm}}=\sigma(\vec{v}_i)$ \algorithmiccomment{Normalizing feature vectors}

\EndFor

\State$\vec{v}=\text {concat}\left\{v_{i}\right\}_{i=1}^{4}$ \algorithmiccomment{Concatenating CSP feature vectors}

\State$\vec{x}_{l}=FR(\vec{v})$ \algorithmiccomment{Feature reduction using ANN}

\State$Label_{pred}=LDA(\vec{x}_{l})$ \algorithmiccomment{Predicting the class labels using LDA classifier}

\end{algorithmic}
\label{alg1}
\end{algorithm}

\subsection{Implementation and Performance Evaluation}
The framework was implemented using PyTorch \cite{paszke2017automatic} and Adam optimization algorithm \cite{kingma2014adam} was used for training the neural networks. For transparency and enabling reproducibility, the code related to the framework and its evaluation is available online\footnote{\url{https://github.com/Singular-Brain/CCSPNet}}. To evaluate the framework’s performance, training and test sets are created as described below:

For the subject-dependent approach, 300 trials belonging to the offline and online portions of session one and the offline portion of session two for each subject are used for training. The online data of session two (100 trials) are used as the test set for each subject.

For the subject-independent approach, the leave-one-subject-out cross-validation (LOSO-CV) method was employed. In LOSO-CV, for each subject, we use the data from other 53 subjects to train a new model. Because of the different nature of the offline (without neurofeedback) and online (with neurofeedback) parts of each session, we used two different training sets, the first one being the online phase data, and the second one, the offline phase data. For a given test subject, each of these training sets consists of 10600 (53×200) trials from other subjects. The test set for each subject is the one used in the SD approach (100 online trials of session two for each test subject). The overall subject-independent accuracy was calculated by averaging over all 54 SI test accuracies obtained by LOSO-CV.

\subsection{Hyperparameters}
There are some structural hyperparameters associated with the CCSPNet framework. The full list of these hyperparameters for both SD and SI approaches are given in Table~\ref{tab2}. The optimal value of each hyperparameter was determined by performing a grid search based on the training loss over a predefined search space.

\begin{table}[h]
\begin{center}
\centering
\caption{CCSPNet Hyperparameters}
\small
\begin{tabular}{cc}
\textbf{Hyperparameter}                                 & \textbf{Value} \\
\hline
\rowcolor[HTML]{F2F2F2} 
Batch size (SI)                                         & 5300           \\
Batch size (SD)                                         & 300            \\
\rowcolor[HTML]{F2F2F2} 
Wavelet parameter update learning rate                  & 0.001          \\
CNN and ANN learning rates                              & 0.01           \\
\rowcolor[HTML]{F2F2F2} 
L1 regularization factor                                & 0.01           \\
L2 regularization factor                                & 0.1            \\
\rowcolor[HTML]{F2F2F2} 
\# Epochs (SI)                                          & 10             \\
\# Epochs (SD)                                          & 20             \\
\rowcolor[HTML]{F2F2F2} 
\# Wavelet kernels                                      & 4              \\
Wavelet kernels frequency range & {[}8, 30{]}    \\
\rowcolor[HTML]{F2F2F2} 
Wavelet kernel length                         & 32             \\

\# Temporal kernels                                     & 4              \\
\rowcolor[HTML]{F2F2F2} 
Temporal kernel length                       & 64             \\
Loss function ratio ($r$)                               & 0.3            \\
       
\end{tabular}
\label{tab2}
\end{center}
\end{table}

\section{Results and Discussion} \label{res}
\subsection{Decoding Accuracy}
Table~\ref{tab3} shows the CCSPNet average decoding accuracy for the SD approach, and the SI approach using the online and offline sets with different batch sizes and number of epochs. All structural hyperparameters in these experiments are set according to Table~\ref{tab2}. The best subject-independent performance was achieved using the offline training set, resulting in an average accuracy of \textbf{74.28\%} over the 54 test subject. The best average accuracy achieved in the subject-dependent approach was \textbf{74.41\%}. For transparency and to enable statistical analysis in future studies, \ref{appendix} provides the detailed subject-by-subject accuracies obtained by the best-performing models, for both SD and SI approaches using the optimum batch size and the number of epochs.

\begin{table}[h]
\begin{center}
\centering
\caption{CCSPNet Classification Results}
\begin{tabular}{@{}cccc@{}}
\begin{tabular}[c]{@{}c@{}}Approach\\(Batch Size, \#Epochs)\end{tabular} & SI (offline) & SI (online) & SD    \\ 
\hline
\rowcolor[HTML]{F2F2F2} 
(2650, 10)                                                                      & 73.39        & 71.74       & -     \\
(2650, 20)                                                                      & 73.83        & 72.57       & -     \\
\rowcolor[HTML]{F2F2F2} 
(5300, 10)                                                                      & \textbf{74.28}        & 72.13       & -     \\
(5300, 20)                                                                      & 73.76        & 73.11       & -     \\
\rowcolor[HTML]{F2F2F2} 
(10600, 10)                                                                     & 73.59        & 71.35       & -     \\
(10600, 20)                                                                     & 73.46        & 71.74       & -     \\
\rowcolor[HTML]{F2F2F2} 
(300, 10)                                                                       & -            & -           & 73.5  \\
(300, 20)                                                                       & -            & -           & \textbf{74.41}
\end{tabular}
\label{tab3}
\end{center}
\end{table}

\subsection{Ablation Experiments}
CCSPNet consists of several components that have been selected and tuned based on multiple experiments and considerations of the BCI literature. However, such a complex structure can be studied more in detail by conducting ablation experiments. The aim of these experiments is to evaluate the importance of each component, and weather any component can be removed without hurting the performance and in order to reduce the number of model parameters. We performed four ablation experiments separately for SD and SI approaches. Considering CSP as the core component of the framework, in each experiment, one of the four other components, WKCNN, TCNN, feature reduction network (FRN), and LDA were removed (in the LDA ablation experiment, a softmax layer was added at the end of FRN for the final classification instead of LDA). The average decoding accuracies and standard deviations of these ablation experiments are given in Table~\ref{tab:ablres}. Comparing only average decoding accuracies shows that the non-ablated model achieves a better performance compared to all ablation experiments. In order to make a more meaningful comparison between the ablated models and the complete model, we performed paired t-tests between the results of each two model in both SD and SI approaches. The null hypothesis is that the mean accuracies of the competing distributions are the same and do not have a significant difference (one model does not outperform the other). The resulting $p$-values of these statistical tests are presented in Tables~\ref{tab:ablsd} and \ref{tab:ablsi}. We reject the null hypothesis for $p$-values that are less than $0.05$. Therefore, based on the tests, the full model outperforms each ablated model significantly. This shows that none of the components can be removed without a loss in performance. In the SD approach, considering the $p$-values when comparing each pair of components, we can observe that removing WKCNN has less impact on performance compared to other components. However, because the $p$-value corresponding to the comparison of the WKCNN-ablated model with the complete model ($0.0460$) showed a significant difference, we did not remove this component from the final SD framework. Other $p$-values in the SD approach indicate that TCNN, FRN, and LDA have the same importance. In the SI approach, by observing the $p$-values comparing FRN with TCNN and LDA, we conclude that FRN plays a less important rule compared to these two components. However, as the $p$-value corresponding to the comparison of the FRN-ablated model with the complete model ($0.0490$) showed a significant difference, we did not remove this component from the final SI framework. Other $p$-values in the SI approach do not show any significant importance difference for other component pairs.

\begin{table}[h] 
\begin{center}
\centering
\caption{Mean accuracy and standard deviation of ablation experiments, removing individual components in the framework, in both SD and SI approaches}
\small
\begin{tabular}{cccc}
\textbf{Approach}         & \textbf{Ablated Component}                                & \textbf{Mean   Accuracy}      & \textbf{Standard   Deviation} \\
\hline
\rowcolor[HTML]{F2F2F2} 
\cellcolor[HTML]{F2F2F2}                                             & WKCNN                                 & 72.61                         & 17.52                         \\
\cellcolor[HTML]{F2F2F2}                                             & TCNN                                 & 70.78                         & 16.37                         \\
\rowcolor[HTML]{F2F2F2} 
\cellcolor[HTML]{F2F2F2}                                             & FRN                                 & 69.98                         & 17.71                         \\
\cellcolor[HTML]{F2F2F2}                                             & LDA                              & 68.91                         & 15.88                         \\
\rowcolor[HTML]{F2F2F2} 
\multirow{-7}{*}{\cellcolor[HTML]{F2F2F2}\textbf{Subject-dependent}} & None                                        & \textbf{74.41}                & 16.75                         \\
                                                                     & WKCNN                          & 70.94                         & 15.81                         \\
                                                                     & \cellcolor[HTML]{F2F2F2}TCNN   & \cellcolor[HTML]{F2F2F2}69.59 & \cellcolor[HTML]{F2F2F2}16.98 \\
                                                                     & FRN                             & 72.57                         & 16.74                \\
                                                                     & \cellcolor[HTML]{F2F2F2}LDA   & \cellcolor[HTML]{F2F2F2}71.39 & \cellcolor[HTML]{F2F2F2}16.71 \\
\multirow{-5}{*}{\textbf{Subject-independent}}                       & None                                        & \textbf{74.28}                & 16.12                        
\end{tabular}
\label{tab:ablres}
\end{center}
\end{table}

\begin{table}[h]
\begin{center}
\centering
\caption{The $p$-values of statistical comparisons in the SD ablation studies}
\begin{tabular}{cccccc}
Ablated component (SD) & None            & WKCNN  & TCNN   & FRN               & LDA              \\ \hline
\rowcolor[HTML]{F2F2F2} 
None      & -                & 0.0460 & 0.001  & \textless{}0.001 & \textless{}0.001 \\
WKCNN     & 0.0460           & -      & 0.0258 & 0.011            & 0.002            \\
\rowcolor[HTML]{F2F2F2} 
TCNN      & 0.001            & 0.0258 & -      & 0.439            & 0.094            \\
FRN        & \textless{}0.001 & 0.011  & 0.439  & -                & 0.351            \\
\rowcolor[HTML]{F2F2F2} 
LDA       & \textless{}0.001 & 0.002  & 0.094  & 0.351            & -     

\end{tabular}
\label{tab:ablsd}
\end{center}
\end{table}

\begin{table}[h] 
\begin{center}
\centering
\caption{The $p$-values of statistical comparisons in the SI ablation studies}
\begin{tabular}{cccccc}
Ablated component (SI) & None             & WKCNN            & TCNN             & FRN    & LDA              \\ \hline
\rowcolor[HTML]{F2F2F2} 
None      & -                & \textless{}0.001 & \textless{}0.001 & 0.049 & \textless{}0.001 \\
WKCNN     & \textless{}0.001 & -                & 0.101            & 0.110 & 0.651            \\
\rowcolor[HTML]{F2F2F2} 
TCNN      & \textless{}0.001 & 0.101            & -                & 0.023 & 0.120            \\
FRN        & 0.049            & 0.110            & 0.023            & -     & 0.098            \\
\rowcolor[HTML]{F2F2F2} 
LDA       & \textless{}0.001 & 0.651            & 0.120            & 0.098 & -               
\end{tabular}
\label{tab:ablsi}
\end{center}
\end{table}

\subsection{Comparative Studies and Statistical Analysis}

Table~\ref{tab4} shows the performance comparison between CCSPNet and other methods on the dataset used for evaluation. In the SD approach, we trained EEGNet \cite{lawhern2018eegnet} using its default hyperparameters and reported its performance as a baseline. The CSP, CSSP, FBCSP, and BSSFO subject-dependent results are reported by Lee et al. \cite{lee2019eeg}. In the SI approach, Pooled CSP, Fused model, and MR FBCSP are reported by Kwon et al. \cite{kwon2019subject}. The results reported by Kwon et al. \cite{kwon2019subject}, Molla et al. \cite{molla2021trial}, and Autthasan et al. \cite{autthasan2021min2net} (the framework knwon as MIN2NET) are also included in our comparisons. In terms of mean accuracy, the CCSPNet has outperformed conventional CSP methods by a large margin and has achieved a state-of-the-are performance comparable to Kwon et al. \cite{kwon2019subject} and Molla et al. \cite{molla2021trial} in the SD paradigm, and comparable to Kwon et al. \cite{kwon2019subject} in the SI paradigm.

In order to make a more meaningful comparison, we performed statistical tests and measure the statistical significance of the results obtained by our framework. Based on the nature of the experiment, the most suitable statistical tests are paired t-test for comparing two models and one-way repeated measures ANOVA for comparing the results obtained by multiple models. The null hypothesis in both paired and multiple statistical tests is that the mean accuracies of the competing distributions are the same. These tests require detailed subject-by-subject data to be available for each method. Since only means and standard deviations are reported by Kwon et al. \cite{kwon2019subject} in their experiments, we can only use the more conservative tests, unpaired t-test, and one-way ANOVA to compare our results with theirs. The detailed results of these statistical tests are reported in Table~\ref{tab5}. In multiple comparisons, one-way ANOVA shows that CCSPNet outperforms the competing methods in SI. In the SD approach, ANOVA does not show a significant difference. One-to-one comparison with other methods shows that our SD mean performance is on a par with the state-of-the-art performance achieved by Kwon et al. \cite{kwon2019subject} and Molla et al. \cite{molla2021trial}, and our SI mean performance is on a par with the state-of-the-art performance achieved by Kwon et al. \cite{kwon2019subject} despite having approximately 10000 fewer parameters (72 million compared to 5036). Furthermore, p-values for one-by-one comparisons show that we outperform all traditional CSP-based methods. We can also compare the performance of CCSPNet on the two different approaches within itself. The result of a paired t-test between the SD approach and the SI approach using the online dataset is (p=0.45). This shows that CCSPNet SI performance is on a par with its SD performance.

\begin{table}[h]
\begin{center}
\centering
\caption{Performance comparison of different SD and SI methods}
\small
\begin{tabular}{cccc}
\textbf{Approach}         & \textbf{Method}                                & \textbf{Mean   Accuracy}      & \textbf{Standard   Deviation} \\
\hline
\rowcolor[HTML]{F2F2F2} 
\cellcolor[HTML]{F2F2F2}                                             & CSP                                 & 68.57                         & 17.57                         \\
\cellcolor[HTML]{F2F2F2}                                             & CSSP                                 & 69.68                         & 18.53                         \\
\rowcolor[HTML]{F2F2F2} 
\cellcolor[HTML]{F2F2F2}                                             & FBCSP                                 & 70.59                         & 18.56                         \\
\cellcolor[HTML]{F2F2F2}                                             & BSSFO                              & 71.02                         & 18.83                         \\
\rowcolor[HTML]{F2F2F2} 
\cellcolor[HTML]{F2F2F2}                                             & EEGNet                                         & 65.31                         & 18.72                         \\
\cellcolor[HTML]{F2F2F2}                                             & MIN2NET                         & 66.06                         & 16.58                \\
\rowcolor[HTML]{F2F2F2} 
\cellcolor[HTML]{F2F2F2}                                             & Molla et al.                                         & 73.85                         & \textbf{15.25}                         \\
\cellcolor[HTML]{F2F2F2}                                             & Kwon et al.                         & 71.32                         & 15.88                \\
\rowcolor[HTML]{F2F2F2} 
\multirow{-7}{*}{\cellcolor[HTML]{F2F2F2}\textbf{Subject-dependent}} & CCSPNet (proposed)                                       & \textbf{74.41}                & 16.75                         \\
                                                                     & Pooled CSP                          & 65.65                         & 16.11                         \\
                                                                     & \cellcolor[HTML]{F2F2F2}Fused model   & \cellcolor[HTML]{F2F2F2}67.37 & \cellcolor[HTML]{F2F2F2}16.01 \\
                                                                     & MR FBCSP                              & 68.59                         & 15.28                \\
                                                                     & \cellcolor[HTML]{F2F2F2}MIN2Net  & \cellcolor[HTML]{F2F2F2}72.03 & \cellcolor[HTML]{F2F2F2}\textbf{14.04} \\
                                                                     & Kwon et al.  &74.15 & 15.83 \\ 
\multirow{-5}{*}{\textbf{Subject-independent}}                       & \cellcolor[HTML]{F2F2F2} CCSPNet (proposed)                                     & \cellcolor[HTML]{F2F2F2} \textbf{74.28}                & \cellcolor[HTML]{F2F2F2} 16.12                        
\end{tabular}
\label{tab4}
\end{center}
\end{table}

\begin{table}[h]
\begin{center}
\centering
\caption{Results of statistical tests comparing CCSPNet performance with other methods}
\small
\resizebox{\columnwidth}{!}{%
\begin{tabular}{c|c|c}
\textbf{Method   (SD)}                                  & \textbf{Paired   t-test (p-value), Unpaired t-test (t-score (106), p-value)} & \textbf{One-way   ANOVA (F-score (8, 477), p-value)}         \\
\hline
\rowcolor[HTML]{F2F2F2} 
\textbf{CSP}                                   & (p\textless{}0.001),   (1.7679, 0.0400)                                      & \cellcolor[HTML]{F2F2F2}                                     \\
\textbf{CSSP}                                  & (0.002), (1.3915,   0.0835)                                                  & \cellcolor[HTML]{F2F2F2}                                     \\
\rowcolor[HTML]{F2F2F2} 
\textbf{FBCSP }                               & (0.007),   (1.1229, 0.1320)                                                  & \cellcolor[HTML]{F2F2F2}                                     \\
\textbf{BSSFO }                               & (0.008), (0.9885,   0.1626)                                                  & \cellcolor[HTML]{F2F2F2}                                     \\
\rowcolor[HTML]{F2F2F2} 
\textbf{EEGNet}                                         & (p\textless{}0.001),   (2.6621, 0.0045)                                      & \cellcolor[HTML]{F2F2F2}                                     \\
\textbf{MIN2NET}                                         & (Data N/A),   (2.6035, 0.0106)                                      & \cellcolor[HTML]{F2F2F2}                                     \\
\rowcolor[HTML]{F2F2F2} 
\textbf{Molla et al.}                                         & (Data N/A),  (0.1817, 0.8562)                                      & \cellcolor[HTML]{F2F2F2}                                     \\
\textbf{Kwon et al.}                         & (Data N/A), (1.9134,   0.292)                                                       & \multirow{-6}{*}{\cellcolor[HTML]{F2F2F2}(1.6945,   0.0972)} \\ \hline
\rowcolor[HTML]{F2F2F2} 
\textbf{Method   (SI)}                                  & \textbf{Unpaired(t-score   (106), p-value)}                                  & \textbf{(F-score   (5, 318) , p-value)}                      \\
\textbf{Pooled   CSP }                          & (2.7827, 0.0032)                                                             &                                                              \\
\cellcolor[HTML]{F2F2F2}\textbf{Fused   model} & \cellcolor[HTML]{F2F2F2}(2.2350,   0.0138)                                   &                                                              \\
\textbf{MR FBCSP}                            & (1.8825, 0.3126)                                                             &                                                              \\
\cellcolor[HTML]{F2F2F2}\textbf{MIN2NET} & \cellcolor[HTML]{F2F2F2}(0.7735,   0.4410)
\\
\textbf{Kwon   et al.} & (0.0423,   0.4832)                                   & \multirow{-4}{*}{(2.9700, 0.0123)}                          
\end{tabular}%
}
\label{tab5}
\end{center}
\end{table}

\subsection{The Effect of Number of Subjects on SI Performance and Robustness}
\begin{figure}
	\centering
	\includegraphics[scale = .35]{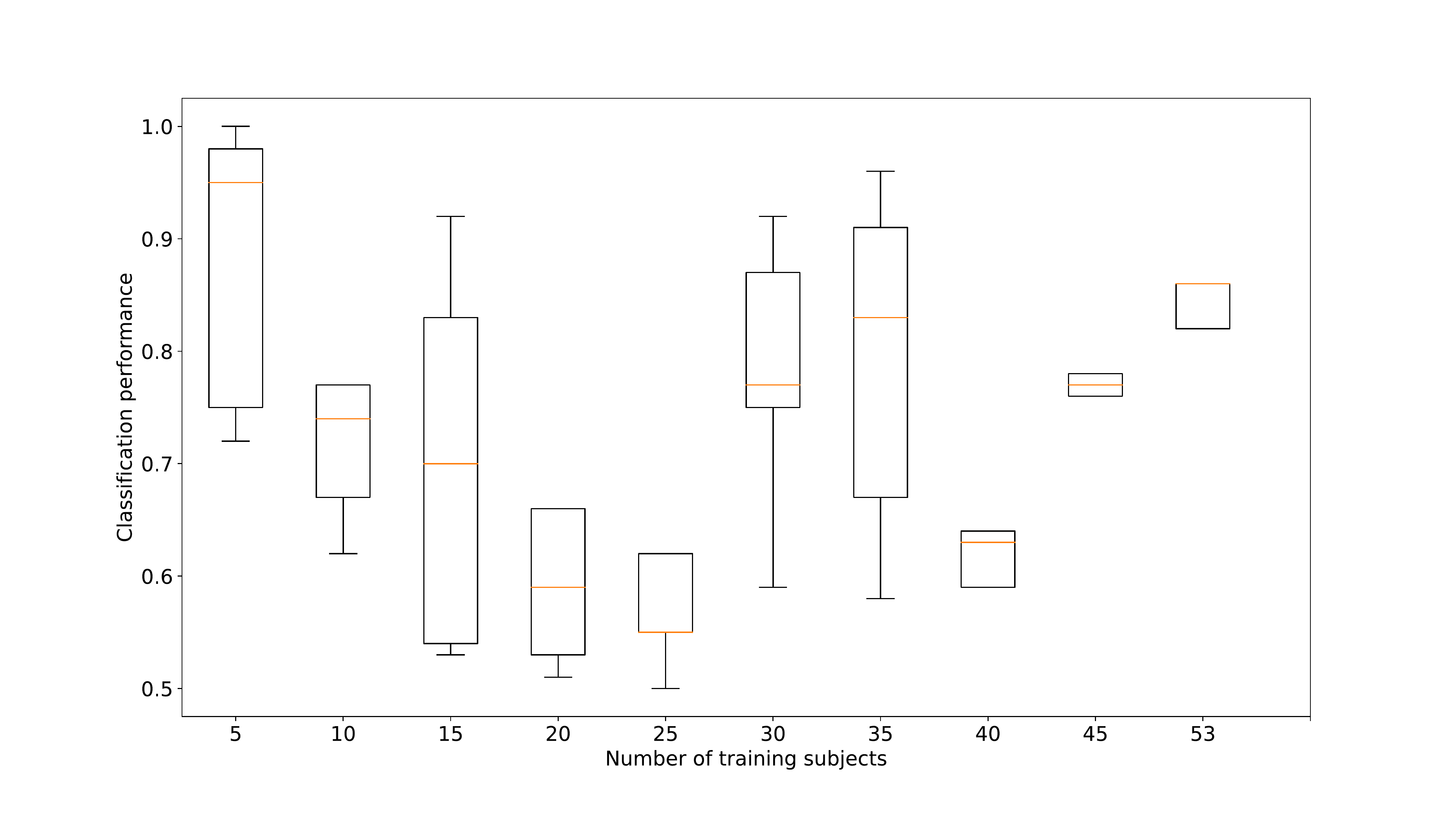}
	\caption{ Effect of the number of training subjects on the CCSPNet subject-independent performance. A specific subject was chosen as the test subject, and the box plots are drawn based on the results of five different training procedures with different random seeds.}
	\label{fig3}
\end{figure}

We analyzed the variations of the subject-independent performance with different number of subjects using an incremental setting. In this setting, we choose one specific test subject and randomly sample a specific number of training subjects and train and evaluate the framework. At each iteration, we increase the number of training subjects. The results of this experiment are given of Figure~\ref{fig3}. The box plots are drawn based on the results obtained in five different training procedures with different random seeds. From this experiment, we can infer that using different subjects for training can either increase or decrease the performance of a subject. This may indicate that different subjects have highly variable brain patterns. When we use a limited number of subjects for training, it may either have a positive effect on the performance if the training subjects share some similar brain patterns with the test subject, or it may worsen the performance if these patterns are radically different from those of the test subject. However, when the number of training subjects increases, as can be seen from the last box plots in Figure~\ref{fig3}, the performance variations become more robust and stable, and the general performance improves. In other words, when there are more subjects present in the training set of the SI approach, we are able to extract features that are generalizable and can create a trade-off between all types of brain patterns.

\subsection{Digital Signal Processing and Deep Learning}
\begin{figure}
	\centering
	\includegraphics[scale=0.4]{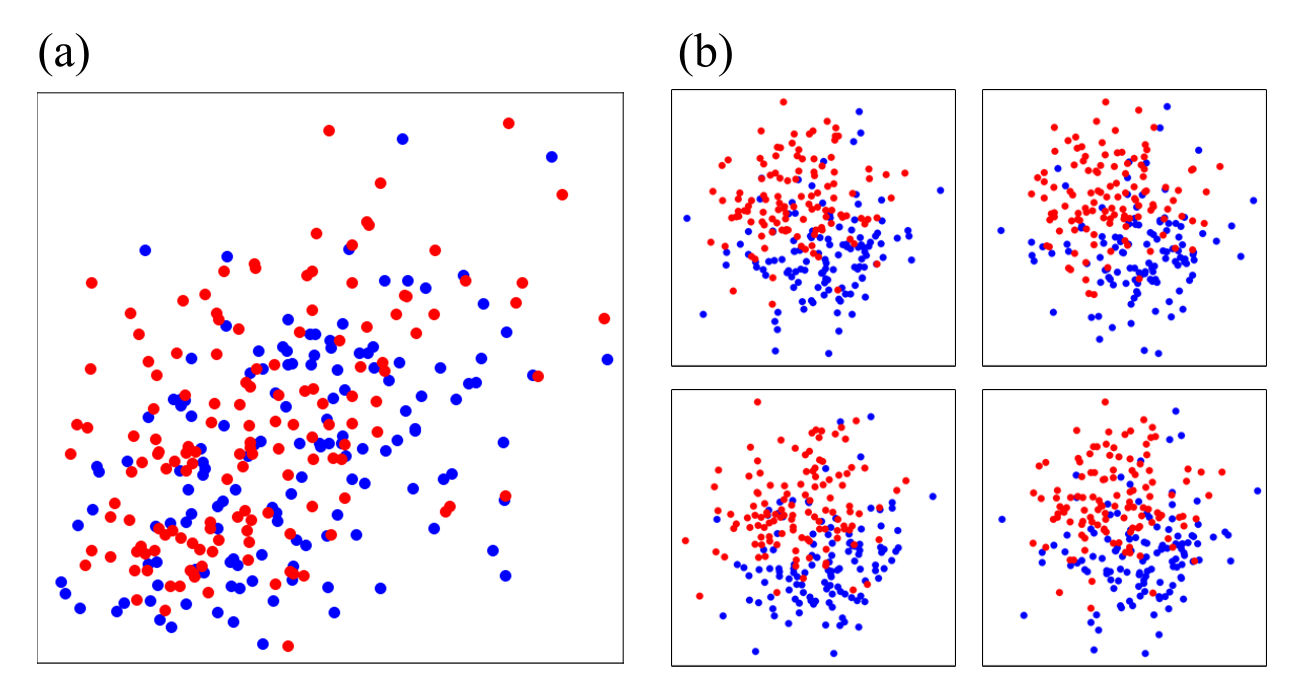}
	\caption{Comparison of the CSP features using normal CSP and CCSPNet. \textbf{a.} CSP feature points (200 points for 100 test trials) for a test subject with a poor performance using normal CSP on the EEG signals. \textbf{b.} The four CSP feature points for the same test subject applied to the outputs of TCNN module of CCSPNet. The red and blue points represent features belonging to each class. Axis $x$ and $y$ in these plots represent the two CSP features respectively.}
	\label{fig4}
\end{figure}

\begin{figure}
	\centering
	\includegraphics[scale = .25]{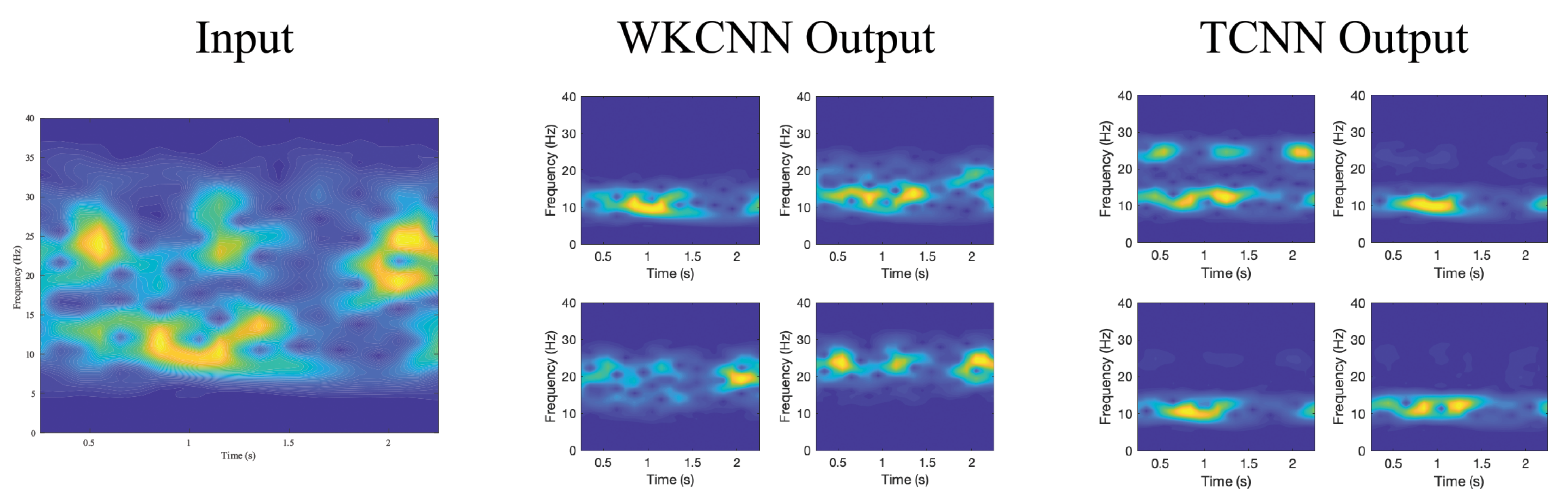}
	\caption{Time-frequency (STFT) representations for a specific subject and EEG channel}
	\label{fig5}
\end{figure}

Deep learning methods have recently shown great success in many tasks involving digital signals. In the BCI literature, as mentioned before, the applications of deep learning have mostly remained limited to the subject-dependent paradigm due to the lack of large-scale datasets that consist of different types of brain patterns. Now that such datasets are becoming available, designing deep learning models that process and classify digital medical signals, such as EEG will become more prevalent, and reliable signal processing techniques will be crucial to design robust and accurate models. In our proposed framework, WKCNN and TCNN layers serve as dynamic digital filters that are optimized to create suitable inputs for the CSP using the novel loss function defined in \eqref{eq12}. TCNN uses normal convolutional kernels that are not represented by a specific mathematical function. These kernels are more flexible and can select a wide range of frequencies to filter the data. Because of this wide frequency coverage, we may call these kernels \emph{non-smooth} filters. On the other hand, wavelet kernels are constructed using an analytical expression and filter the data to a specific frequency range. Therefore, we may call these kernels \emph{smooth} filters. Figure~\ref{fig5} shows three time-frequency plots, using short-time Fourier transform (STFT), for a specific subject and EEG channel. These contours are plotted before any pre-processing (the left image), after applying WKCNN, and after TCNN. As illustrated in this figure, first, the WKCNN discovers features in specific frequency ranges (the middle image). Afterwards, the more flexible TCNN combines features from different frequency bins detected by the wavelet kernels (the right image). As a result of these dynamic filters, CSP can achieve much better performance compared to raw inputs. Traditional CSP uses filtered signals from a single frequency band. A more advanced algorithm such as Filter Bank CSP (FBCSP) \cite{lotte2009comparison} uses multiple frequency bands and applies the CSP algorithm on all of the band-filtered signals. However, these frequency bands are still constant and predetermined, and therefore the model may fail to capture spectral information from more diverse frequency bins. Our dynamic filtering solves this problem and improves the quality of CSP inputs. This improvement is shown in Figure ~\ref{fig4} where our CSP outputs are compared with the one applied to raw signals. The left scatter plot in Figure~\ref{fig4} shows the normal CSP output of a test subject (200 feature points for 100 test trials). The red and blue dots indicate the feature points of each class. As it may be predicted, normal CSP fails to distinguish between two classes and yields a poor performance for this subject. On the other hand, the right plot shows the proposed framework CSP features for each of the four CNN temporal kernels, for the same test subject. In this case, CSP outputs show more separation between classes, and the subsequent decoding performance improved significantly compared to the model based on normal CSP. The same conclusion can also be drawn by comparing the accuracies provided in Table~\ref{tab4}.

A major factor in achieving a better performance compared to normal CSP methods may be the use of the loss function in \eqref{eq12}. By minimizing this loss, we are trying to make the CSP feature vector closer to the optimal target vector corresponding to the class label. Since the CSP inputs are generated by the CNN, by using this loss as a feedback mechanism to update the CNN weights, we are indirectly improving the CSP features by providing it with more refined inputs.
 
\subsection{Trade-off between SD and SI performance}

\begin{figure}[htp]
	\centering
	\includegraphics[scale=0.4]{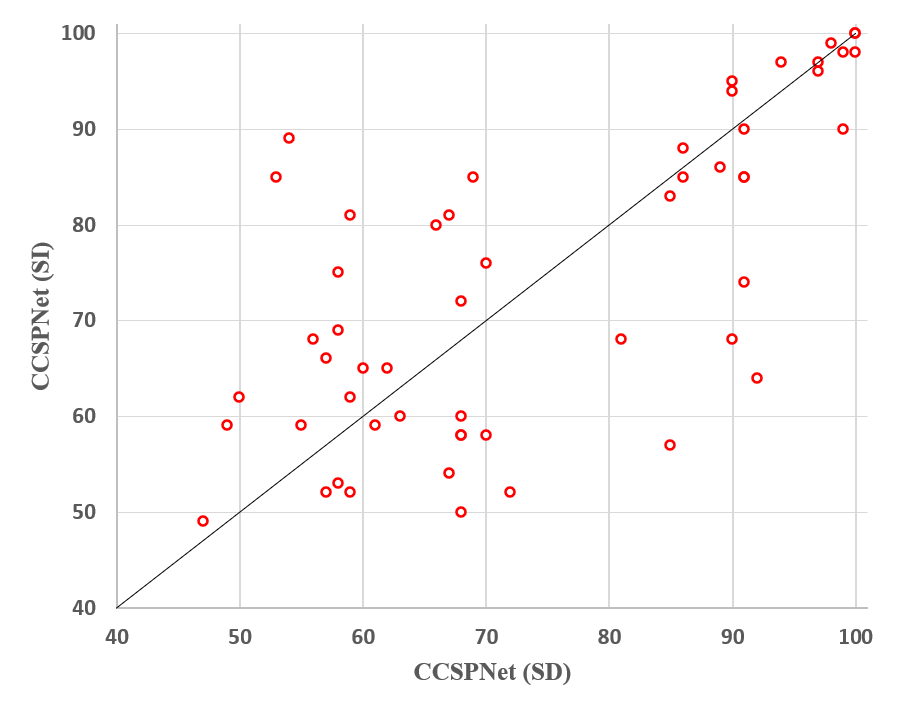}
	\caption{Subject-by-subject SD and SI Performance of the Proposed Framework}
	\label{fig6}
\end{figure}

Figure~\ref{fig6} shows a scatter plot comparing the subject-by-subject SI and SD performance of the proposed framework. Each point represents a subject, and the horizontal and vertical axes are representing SD and SI performances, respectively. We observe that the SI model has increased the classification accuracy of many test subjects with poor and mediocre performance in the SD approach, while decreasing the accuracy of many subjects with a high SD performance. The reason behind this phenomenon may be that by utilizing a large number of available trials and the dynamic WKCNN-TCNN filtering, we are able to extract generalizable features applicable to the brain signals of a wide variety of subjects, which in turn may improve the performance of subjects with low or average SD performance who have no inherent brain signal patterns. On the other hand, as hypothesized by Kwon et al. \cite{jayaram2016transfer}, subjects with excellent subject-dependent accuracy may have their own specific and inherent EEG signal features that are responsible for their excellent SD performance. Therefore, applying a model with generalizable features to their brain signals may result in a performance deterioration. This reduction in excellent subjects classification accuracy is not a concern, because in a potential application, after detecting that a subject has inherent brain patterns and a good SD performance, the mode of the classifier can be set to subject-dependent. However, this is not possible for users with no inherent motor-imagery patterns, and the only plausible solution for these subjects is to improve their subject-independent accuracy by using generalizable features. Consequently, improving the poor and average-performing subjects in the SI approach is a great achievement for a BCI model.

\subsection{Limitations and Further works}
As the title of the paper suggests, we believe that our work can be a step towards real-world BCI applications, and we base our claim on the subject-independence and the compactness of our framework. The following limitations and possible extensions are worth considering. The proposed framework can be extended and applied to multi-class EEG datasets. Since normal CSP is applicable to two-class datasets, this task requires using a multi-class CSP and a multi-class classifier at the end \cite{grosse2008multiclass}. This modification is not done in the current work because our focus have been on the subject-independent paradigm that requires a large-scale EEG dataset such as the one provided by Lee et al. \cite{lee2019eeg}. If similar large-scale datasets become available for multi-class BCI tasks, CCSPNet can be easily modified and applied to them for both SD and SI classification. Otherwise, normal subject-dependent experiments may be done on multi-class datasets after the aforementioned modification. Availability of such datasets would open the door to BCI models for complex motor imagery tasks that could be used in real-world rehabilitation systems in the future. Furthermore, possible changes may be done to the proposed architecture to improve its performance. For example, other types of classifiers can be applied in the last stage of the framework, or as done by Ang et al. \cite{ang2008filter}, the LDA may be applied only for class discrimination, and another classifier can be used for the final classification. Additionally, although it could hurt compactness, we may add more layers and channels to the spectral feature extractors and create a larger model. Finally, to be applicable to a real-world application, a potential application would require a robust and highly accurate model which requires gathering even larger EEG databases. Such a dataset can be used by a real-world model that is built upon the current BCI frameworks, such as CCSPNet.

\section{Conclusion} \label{con}
In this work, we proposed CCSPNet, a compact subject-independent framework designed for decoding motor imagery BCI tasks. This hybrid framework takes advantage of deep learning along with a novel loss function and feedback mechanism to generate spectral features for CSP, and to reduce its features before applying an LDA classifier. The compactness and the state-of-the-art performance achieved together by our framework is, to this date, scarce in motor imagery, and generally in the BCI literature. However, we hope that the use of similar compact and accurate frameworks with novel approaches and architectures will become more prevalent in experimental and real-world BCI.

\hfill \break
\textbf{We have no Conflict of Interest}

\pagebreak
\appendix
\section{Subject-by-subject Results} \label{appendix}

The detailed subject-by-subject performance of CCSPNet is given in the following table.

\begin{table}[htp]
\begin{center}
\centering
\caption{Extended CCSPNet Results}
\small
\scalebox{0.36}{
\resizebox{\columnwidth}{!}{%
\begin{tabular}{ccc}
\textbf{\begin{tabular}[c]{@{}l@{}}\backslashbox{Approach}{Subject}\end{tabular}} & \textbf{SD}                   & \textbf{SI}                      \\
\hline
\cellcolor[HTML]{F2F2F2}\textbf{1}                                        & \cellcolor[HTML]{F2F2F2}91    & \cellcolor[HTML]{F2F2F2}85      \\
\textbf{2}                                                                & 91                            & 74                              \\
\cellcolor[HTML]{F2F2F2}\textbf{3}                                        & \cellcolor[HTML]{F2F2F2}97    & \cellcolor[HTML]{F2F2F2}96       \\
\textbf{4}                                                                & 68                            & 58                              \\
\cellcolor[HTML]{F2F2F2}\textbf{5}                                        & \cellcolor[HTML]{F2F2F2}86    & \cellcolor[HTML]{F2F2F2}88     \\
\textbf{6}                                                                & 91                            & 85                               \\
\cellcolor[HTML]{F2F2F2}\textbf{7}                                        & \cellcolor[HTML]{F2F2F2}81    & \cellcolor[HTML]{F2F2F2}68     \\
\textbf{8}                                                                & 66                            & 80                               \\
\cellcolor[HTML]{F2F2F2}\textbf{9}                                        & \cellcolor[HTML]{F2F2F2}70    & \cellcolor[HTML]{F2F2F2}76      \\
\textbf{10}                                                               & 68                            & 58                              \\
\cellcolor[HTML]{F2F2F2}\textbf{11}                                       & \cellcolor[HTML]{F2F2F2}47    & \cellcolor[HTML]{F2F2F2}49       \\
\textbf{12}                                                               & 58                            & 53                               \\
\cellcolor[HTML]{F2F2F2}\textbf{13}                                       & \cellcolor[HTML]{F2F2F2}60    & \cellcolor[HTML]{F2F2F2}65       \\
\textbf{14}                                                               & 57                            & 66                               \\
\cellcolor[HTML]{F2F2F2}\textbf{15}                                       & \cellcolor[HTML]{F2F2F2}59    & \cellcolor[HTML]{F2F2F2}62       \\
\textbf{16}                                                               & 58                            & 69                               \\
\cellcolor[HTML]{F2F2F2}\textbf{17}                                       & \cellcolor[HTML]{F2F2F2}67    & \cellcolor[HTML]{F2F2F2}81       \\
\textbf{18}                                                               & 94                            & 97                              \\
\cellcolor[HTML]{F2F2F2}\textbf{19}                                       & \cellcolor[HTML]{F2F2F2}85    & \cellcolor[HTML]{F2F2F2}83       \\
\textbf{20}                                                               & 59                            & 81                               \\
\cellcolor[HTML]{F2F2F2}\textbf{21}                                       & \cellcolor[HTML]{F2F2F2}100   & \cellcolor[HTML]{F2F2F2}98      \\
\textbf{22}                                                               & 90                            & 95                               \\
\cellcolor[HTML]{F2F2F2}\textbf{23}                                       & \cellcolor[HTML]{F2F2F2}62    & \cellcolor[HTML]{F2F2F2}65      \\
\textbf{24}                                                               & 49                            & 59                               \\
\cellcolor[HTML]{F2F2F2}\textbf{25}                                       & \cellcolor[HTML]{F2F2F2}92    & \cellcolor[HTML]{F2F2F2}64      \\
\textbf{26}                                                               & 55                            & 59                               \\
\cellcolor[HTML]{F2F2F2}\textbf{27}                                       & \cellcolor[HTML]{F2F2F2}50    & \cellcolor[HTML]{F2F2F2}62      \\
\textbf{28}                                                               & 100                           & 100                              \\
\cellcolor[HTML]{F2F2F2}\textbf{29}                                       & \cellcolor[HTML]{F2F2F2}99    & \cellcolor[HTML]{F2F2F2}90      \\
\textbf{30}                                                               & 68                            & 60                               \\
\cellcolor[HTML]{F2F2F2}\textbf{31}                                       & \cellcolor[HTML]{F2F2F2}68    & \cellcolor[HTML]{F2F2F2}72      \\
\textbf{32}                                                               & 98                            & 99                               \\
\cellcolor[HTML]{F2F2F2}\textbf{33}                                       & \cellcolor[HTML]{F2F2F2}99    & \cellcolor[HTML]{F2F2F2}98      \\
\textbf{34}                                                               & 68                            & 50                               \\
\cellcolor[HTML]{F2F2F2}\textbf{35}                                       & \cellcolor[HTML]{F2F2F2}85    & \cellcolor[HTML]{F2F2F2}57      \\
\textbf{36}                                                               & 91                            & 90                               \\
\cellcolor[HTML]{F2F2F2}\textbf{37}                                       & \cellcolor[HTML]{F2F2F2}90    & \cellcolor[HTML]{F2F2F2}94        \\
\textbf{38}                                                               & 59                            & 52                                \\
\cellcolor[HTML]{F2F2F2}\textbf{39}                                       & \cellcolor[HTML]{F2F2F2}69    & \cellcolor[HTML]{F2F2F2}85      \\
\textbf{40}                                                               & 70                            & 58                                \\
\cellcolor[HTML]{F2F2F2}\textbf{41}                                       & \cellcolor[HTML]{F2F2F2}57    & \cellcolor[HTML]{F2F2F2}52       \\
\textbf{42}                                                               & 58                            & 75                               \\
\cellcolor[HTML]{F2F2F2}\textbf{43}                                       & \cellcolor[HTML]{F2F2F2}90    & \cellcolor[HTML]{F2F2F2}68        \\
\textbf{44}                                                               & 100                           & 100                              \\
\cellcolor[HTML]{F2F2F2}\textbf{45}                                       & \cellcolor[HTML]{F2F2F2}97    & \cellcolor[HTML]{F2F2F2}97        \\
\textbf{46}                                                               & 89                            & 86                               \\
\cellcolor[HTML]{F2F2F2}\textbf{47}                                       & \cellcolor[HTML]{F2F2F2}54    & \cellcolor[HTML]{F2F2F2}89        \\
\textbf{48}                                                               & 67                            & 54                                \\
\cellcolor[HTML]{F2F2F2}\textbf{49}                                       & \cellcolor[HTML]{F2F2F2}63    & \cellcolor[HTML]{F2F2F2}60        \\
\textbf{50}                                                               & 72                            & 52                               \\
\cellcolor[HTML]{F2F2F2}\textbf{51}                                       & \cellcolor[HTML]{F2F2F2}53    & \cellcolor[HTML]{F2F2F2}85       \\
\textbf{52}                                                               & 86                            & 85                               \\
\cellcolor[HTML]{F2F2F2}\textbf{53}                                       & \cellcolor[HTML]{F2F2F2}61    & \cellcolor[HTML]{F2F2F2}59        \\
\textbf{54}                                                               & 56                            & 68                                \\
\cellcolor[HTML]{F2F2F2}\textbf{Mean}                                     & \cellcolor[HTML]{F2F2F2}74.41 & \cellcolor[HTML]{F2F2F2}74.28     \\
\textbf{Standard   Deviation}                                             & 16.75                         & 16.12                             \\
\cellcolor[HTML]{F2F2F2}\textbf{Median}                                   & \cellcolor[HTML]{F2F2F2}68.5  & \cellcolor[HTML]{F2F2F2}73        \\
\textbf{Range}                                                            & 53 (47-100)                            & 51 (49-100)                               
\end{tabular}%
}
}
\end{center}
\label{taba1}
\end{table}

\pagebreak
\bibliography{bib}

\end{document}